# Confucius, Cyberpunk and Mr. Science:
Comparing AI ethics between China and the EU

**Pascale Fung** and **Hubert Etienne**
Centre for Artificial Intelligence Research (CAiRE)
The Hong Kong University of Science and Technology

1. **Introduction**

The exponential development and application of artificial intelligence triggered an unprecedented global concern for potential social and ethical issues. Stakeholders from different industries, international foundations, governmental organisations and standards institutions quickly improvised and created various codes of ethics attempting to regulate AI. This exercise, in direct reaction to a perceived great challenge, is only comparable to that of controlling the proliferation of nuclear weapons in the 1960's. Its objective is no less than to reach a global agreement on common ethical standards to regulate one of the most promising technologies whose crucial strategic implications on both business and political grounds are already well acknowledged. The resulting profusion of documents on AI ethical standards, as much as 84 identified by Jobin et al. 2019 [1] and 160 by the AI Ethics guidelines Global Inventory [2], however, deserve to be scrutinized.

A major concern is the large homogeneity and presumed consensualism around these principles. Jobin et al. identified 11 clusters of ethical principles among 84 documents (2019) and Fjeld et al [3] found 8 key themes across 36 of the most influential of these (2020). They both noted a general convergence, which leads Fjeld et al. to conclude that "the conversation around principled AI is beginning to converge" and that "these themes may represent the 'normative core' of a principle-based approach to AI ethics and governance" [5]. However, we argue that ethics, by nature, is not consensual. While it is true that some ethical doctrines, such as the famous Kantian deontology, aspire to universalism, they are however not universal in practice. In fact, ethical pluralism is more about differences in which relevant questions to ask rather than different answers to a common question. When people abide by different moral doctrines, they tend to disagree on the very approach to an issue. Even when people from different cultures happen to agree on a set of common principles, it does not necessarily mean that they share the same understanding of these concepts and what they entail.

In order to better understand the philosophical roots and cultural context underlying ethical principles in AI, we propose to analyse and compare the ethical principles endorsed by the Chinese National New Generation Artificial Intelligence Governance Professional Committee (CNNGAIGPC) [4] and those elaborated by the European High-level Expert Group on AI (HLEGAI) [5]. China and the EU have very different political systems and diverge in their cultural heritages. In our analysis, we wish to highlight that principles that seem similar *a priori* may actually have different meanings, derived from different approaches and reflect distinct goals.

| Chinese ethical principles | E.U. key requirements |
|---|---|
| 1. Harmony and friendship. | Societal and environmental well-being |
| 2. Fairness and justice. | Diversity, non-discrimination and fairness |
| 3. Tolerance and sharing. | Human agency and oversight |
| 4. Respect privacy. | Privacy and data governance |
| 5. Safe and controllable. | Technical Robustness and safety |
| 6. Share responsibilities. | Transparency |
| 7. Open collaboration. | Accountability |
| 8. Agile governance. | |

Table 1. The ethical principles endorsed by the Chinese National New Generation Artificial Intelligence Governance Professional Committee (CNNGAIGPC) [6] and those elaborated by the European High-level Expert Group on AI (HLEGAI)

## 2. Promotional vs Prohibitive Roles of the Government

At first glance, the Chinese ethical principles seem similar to those of the E.U. in many aspects. Both notably promote fairness, robustness, privacy, safety and transparency. Their prescriptive approaches, however, reveal different cultural perspectives associated with different objectives.

*2.1 Confucianism vs. Enlightenment: A Collective vs Individualist Cultural Heritage*

Confucian philosophy has shaped the governing system in China and the rest of East Asia for centuries. It emphasizes the "rule for the people", rather than "rule by the people", and favours an elitist leadership, associating political mandates with competence and merit. The Chinese government's belief in "doing the right thing" for its citizens is informed by the Confucian ideas of virtuous authority and exemplary person, grounded in *ren* (humaneness), *yi* (appropriateness), *li* (rite), and *zhi* (wisdom). This Confucian philosophical tradition explains the community-focused and goal-oriented perspective, from which the Chinese guidelines derive, together with the promotions of principles such as "harmony and friendship", "shared responsibilities", "tolerance and sharing", and "open collaboration".

"A high sense of social responsibility and self-discipline" is also expected from individuals to harmoniously partake in a community while promoting tolerance, shared responsibilities and open collaboration. The emphasis is explicitly informed by the Confucian value of 'harmony' as an ideal balance to be achieved through the control of extreme passions to avoid conflicts. Other than a stern admonition against 'illegal use of personal data', such value leaves little room for constraining rules. These principles are not paths to regulation. Rather, they are framed to guide AI developers in the 'right way' for the collective good of the society.

The European ethics principles, in contrast, emerge from a more individual-focused and rights-based approach. They express a different aspiration, rooted in the Enlightenment tradition, and coloured by the European history. Their primary goal is to protect individuals against well-identified harms. Whereas the Chinese principles emphasize the promotion of good practices, the E.U. focuses on the prevention of evil consequences. The former draws a direction for the development of AI, so that it contributes to the improvement of society. The latter sets limitations to its uses, so that it does not happen at the expense of certain categories of people.

This distinction is clearly illustrated by the presentation of fairness, diversity and inclusiveness. While the E.U. emphasizes fairness and diversity with regard to individuals from specific demographic groups (specifying gender, ethnicity, disability, etc.), Chinese guidelines urge for the upgrade of 'all industries', reduction of 'regional disparities' and prevention of data monopoly. While the E.U. insists on the protection of vulnerable *persons* and potential victims, China prescribes 'inclusive development through better education and training, support'.

The individualist perspective reflected by the European approach to AI ethics should, however, not be mistaken for a form of selfish moral individualism, but rather as a result from the European history of individual reasoning. It is worth noting that the first claim of the Enlightenment did not target political self-determination nor the possibility for people to partake in collective decision-making, but rather ontological autonomy, freeing them from the subjection to the king and the power to the state. This was famously defined by E. Kant as "man's emergence from his self-incurred state of immaturity".

This point is also illustrated in the *Declaration of Human and Citizen Rights* which gives prevalence of a citizen's protection against political power abuse, i.e. negative rights, over positive rights. Finally, the repeated clashes of European nationalism that culminated in WWII and the trauma of Totalitarianism acted as a powerful reminder to the Europeans of the dangers of political holism. Consequently, European societies have shown a clear preference of individualist and rights-based approaches of governance.

*2.2 Promotion vs Prohibition*

AI governance in both the EU and China are led by the transnational and national governments, in consultation with industry stakeholders and academic experts. It is therefore pertinent to compare the actual and perceived roles of their governments in setting AI ethical guidelines. Philosophers have debated the compatibility of Confucian values with Western liberal democracies. There have also been debates on normative versus empirical legitimacy of a government, where scholars study the question of why "the observed level of regime legitimacy under non-democratic regimes has been substantially higher than either established or emerging democracies" [7].

Based on the findings of the Asian Barometer Survey, Doh Chull Shin found that the majority of East Asians in countries with a Confucian tradition tend to be attached to "paternalistic meritocracy", prioritize economic well-being over freedom, and define democracy in substantive (rather than procedural) terms." [8] China is more of an assumed authoritarian technocracy than an anti-democracy. Its political elite, mainly promoted by merit and composed of civil servants mostly with backgrounds in science, technology, engineering and mathematics (STEM), has adopted a pragmatic approach to AI ethics, grounded in existing applications and driven by the needs of the society.

The Chinese leadership at each level, all the way to the highest office, routinely holds "workshops" with academicians and scientists to learn about the latest trends in advanced technologies, such as AI, quantum computing, or biotechnologies. This proximity between the political leaders and the scientific research, together with the greater control exercised by the government on the development of this technology, are included in the centralized planning of the economy to serve national strategic objectives. It explains why the Chinese approach to AI ethics is neither ideological, nor speculative. It does not approach AI governance with the question of what AI might become, but rather what it is doing , and prevents the harm that can come from it being exploited in the foreseeable future.

The Chinese government is not averse to regulations, but in the case of AI governance, it is unlikely to regulate with a broad brush before AI has been widely applied and has found to be of serious negative impact in specific areas or posing danger to the society. Nevertheless, in areas of immediate societal impact and concern, such as data privacy, China has also devised strict laws and regulations. Many have found the Chinese regulation of personal data privacy to be the most comprehensive and robust in Asia, precisely because it has the largest digital market in the region. Even though the guideline calls for mere "respect for privacy", it is understood that companies must exercise extreme self discipline in terms of user data protection.

In contrast, the traditional training of the European political elite does not always allow the same interest, nor understanding, for AI and new technologies in general. Furthermore, the experience with Totalitarianism in Europe always serves as a reminder, calling for prudence. It also explains the greater scepticism from the European citizens about technologies, especially those that can be used for surveillance purposes. A recent example is the strong general reluctance of the French population to adopt the official covid-tracking application, "Tous Anti-Covid", out of fear of what it could be used for by the government. This is why the European approach to AI ethics since the beginning was conceived as both a way to regulate the private sector from foreseeable risks, and to prevent systematic distrust against AI by the public. It is intended to provide some sort of guarantee against potential abuses from public-private partnerships between governments and AI companies.

The different historical contexts in China and the E.U. mentioned above translate into two different types of moral imperatives. The European requirements, centred on satisfying initial conditions, dictate a strict abidance by deontologist rules in the pure Kantian tradition. In contrast, the Chinese principles, referring to an ideal to aim for, express rather softer constraints at different levels, as part of a process to improve society. For the Europeans the development of AI *'must* be fair', for the Chinese it should 'eliminate prejudices and discriminations *as much as possible'*. The EU 'requires processes to be transparent', China's requires to "continuously *improve'* transparency. The E. U. principles aim to protect European citizens from vertical and horizontal abuses, conscient of the danger of nationalism. The Chinese governance system, in contrast, adopts a holistic approach, holding that the social group it forms is not to be reduced to the sum of its parts, but produces something more, namely the Chinese nation. Its ethical principles thus aim to benefit Chinese citizens through the service of the Chinese nation, considered as a common good citizens are associated with.

3. **Utopian vs Dystopian Vision of the General Public**

Other than roles of the governments, the two ethical guidelines are informed by opposing views from the European and Chinese public regarding AI. The main fears expressed by western society towards AI are related to privacy and surveillance, job automation [9], and the possibility of a loss of control resulting in existential risks for humanity [10]. These are greatly dependent on people's trust in political and technology leadership, on the narratives surrounding the development of AI in mainstream media, and on the representation of AI in science fiction.

*3.1 Trust in the Government*

Public opinion studies show that Chinese people are largely supportive of AI, which they associate with a great potential to benefit society, and as an engine of economic growth. Strong government support, a vibrant commercial market for AI, and media content favorable to AI all contribute to this positive perception [11]. A comparative study of German, Chinese and UK participants to assess the Attitude Towards Artificial Intelligence (ATAI scale) showed that the Chinese scored the highest on the ATAI Acceptance scale and lowest on the ATAI Fear scale [12]. Around 70 percent of Chinese respondents stated that they trust artificial intelligence, according to another survey conducted by Ipsos in September 2018 [13].

Overall, Asian public opinion tends to be more favourable to AI. Fo example, a Pew Research Center survey in 2020 found that "majorities in most Asian public – Singapore (72%), South Korea (69%), India (67%), Taiwan (66%) and Japan (65%) – surveyed see AI as a good thing" for society, whereas more than half of the EU population view AI as negative. "In France, for example, views are particularly negative: Just 37% say AI has been good for society, compared with 47% who say it has been bad for society. In the U.S. and UK, about as many say it has been a good thing for society as a bad thing." [14]

The European historical context somewhat led to a general state of distrust in governments in many liberal democracies, which is described as the "counter-democracy" by (Rosanvallon 2006) [15]. In France [16], as in the US [17], For instance, more than three quarters of the citizens think their political representatives behave unethically. The fear of AI being used by governments for mass surveillance is a major concern, and public-private collaborations are also regarded with high scepticism. From the private sector, multiple incidents and scandals related to user privacy, surveillance and nudging involving top technology companies in the past few years severely dampened consumer enthusiasm, together with the perception of these companies' intentions to do good or to operate responsibly [18].

This trust gap is particularly well illustrated by the perception of 'privacy'. Data privacy is promoted by both the European and the Chinese ethical guidelines, but with different meanings. The European promotion of privacy, as highlighted by GDPR, encompasses the protection of individual data from both state and commercial entities. The Chinese privacy guidelines in contrast only targets private companies and potential malicious agents. Whereas personal data is strictly protected both in the EU and in China from commercial entities, the state retains full access in China. Such a practice would be shocking in Western countries; it is however readily accepted by Chinese citizens, accustomed to living in a protected society and have consistently shown high trust in their government [19].It is within the social norm in China where Chinese parents routinely have access to their children's personal information to provide guidance and protection. This difference goes back to the Confucian tradition of trusting and respecting the heads of state and family. The trust is nowadays strengthened by the great economic growth the country's leaders succeeded in achieving. A recent survey showed that Chinese government's successful domestic management of the Covid-19 crisis is likely to inspire more trust by its citizens [20].

The trust gap is therefore related to the perception of government competency, and thus to the objectives these governments aim to achieve with AI. The most developed European countries are former global powers, which gave up on their past expansionist ambitions, and now focus on domestic policies to solve their social issues, while trying not to be left behind in the innovation race. In contrast, China has recently established itself as a world leading economy. This rapid ascend onto the world stage, together with the clear ambition to challenge the U.S. leadership has won the trust from the Chinese people in the actions of their government, including the strategic support given to AI.

*3.2. Media Narrative*

Another source of distrust in AI is due to the large volume of sensationalism, disinformation, misinformation and conspiracy theories surrounding the technology. Identification of disinformation, misinformation and mal-information has become a pressing challenge and a main responsibility of internet companies, and government organizations around the world, especially following the revelation that a vast misinformation campaign served to influence the American election in 2016.

Disinformation about Covid-19 and vaccines are considered harmful and are flagged by social media sites. UNESCO published "Journalism, 'Fake News' and Disinformation: A Handbook for Journalism Education and Training" to address the challenge. In Asia, Singapore has outlawed fake news under its *Protection from Online Falsehoods and Manipulation Bill*. This bill requires service providers to remove content deemed false or allows the government to block it. Disinformation is equally outlawed in China. Such paternalistic control of (dis)information exists in Confucian societies precisely because Confucian philosophy mandates that the legitimacy of a government is measured by its "wisdom" and its merit - rule for the people. This is one area in which the regulation is more strict than in Europe. Disinformation, whether it is about Covid-19 or AI, is seen to germinate and promote discord in a society and is therefore not tolerated in a Confucian society.

It is also found that media in China speak mostly favourably of AI, following a top-down national AI strategy. In contrast, the media and press in the West are more inclined to report from a negative angle, as journalists mostly see themselves as critics of Big Tech and AI. While journalism represents a valuable safeguard to reveal and inform people about AI risks, journalists often cede to the temptation to play on fears to attract readership – which are often greatly inflated and remain persistent even after they had been debunked by academics [21]. Misrepresentations and half-truths of the technology persist and continue to contribute to some of the unfounded fears and hysteria about AI. It compelled many AI researchers to provide explainers such as "The Seven Deadly Sins of AI Predictions" by Rodney Brooks [22], who wrote that "mistaken predictions lead to fears of things that are not going to happen", and explained how a general lack of understanding of how machines work contribute to magical thinking and the hysteria around AI.

*3.3. Cyberpunk Culture*

The gap in the cultural representation of AI, perceived as a force for good in China, and as a menacing force in the dystopian technological future in the Western world, is also enhanced by the influence of popular culture. Robots are assistants and companions in the utopian Chinese vision, they tend to become insurrectional machines as portrayed by a Western media heavily influenced by the cyberpunk subgenre of sci-fi, embodied by The Blade Runner, The Matrix and the Black Mirror series.

Cyberpunk emerged in the 1960s in the West, as a subgenre of science fiction. It represents a view of a high-tech future where social orders are broken down and renegade rebel forces battle against a Big Brother government that uses technology to control the people. This vision, embodied in the works of Philip Dick et al., is a stark departure from the Utopian visions of a technological future espoused by Isaac Asimov and Jules Verne in previous generations of science fiction.

Scholars have also found that public opinions can be shaped by popular culture. In particular, (Young Carpenter 2018) found that "consumption of frightening armed AI films is associated with greater opposition to autonomous weapons" [23]. Since there is no autonomous weapons system in existence today, the only impression people associate with such systems and futuristic AI is from sci-fi literature and films, "Sci-fi as a genre, and certain iconic killer robot films in particular, appears most salient in rhetorical arguments against such weapons. […And] robopocalyptic films themselves have been likelier to encourage a cautionary rather than techno-optimistic sentiment on armed AI, among at least sci-fi literate members of the American public."

Chinese science fiction in the early decades of the 20th century was mostly translated from Soviet literature and mostly for children. (The first author was impressed by the science fiction story of service robots and computer shopping in a children's book when she was a child in the 1970s and decided to become a robotist in the future.) Whereas in Chinese literature there is no tradition of description of a utopian future there is not exactly a dystopian and cyberpunk influence either. Mainland China was mostly closed to the outside world before the 1980s, therefore was not influenced by the rise of the cyberpunk culture nor dystopian visions of either George Orwell's "1984" nor that of 2001: A Space Odyssey. The influence of the Soviet Union also stopped in 1960s when the diplomatic relationship between China and the USSR was severed, preventing the Chinese public from the dark visions of Stanisław Lem, for example.

It is interesting to note an opposite trend to that of China in a country with a similar Buddhist/Confucian culture: the Japanese are found to have a relatively low level of trust in their government, in particular following the Fukushima nuclear plant crisis in 2011. Their trust in the government ranks below that of many EU countries including Germany [24]. Despite being a world leading pioneer in robotics, and despite having in general a technophile population who love the latest gadgets, there is a general doomsday malaise from a collective memory of the only atomic bomb detonation in history. The cyberpunk animation classic Akira, produced in 1988, foretold a post-apocalyptic dystopian future in 2019 rife with anti-government protests and gang violence, superpowers and government sponsored assassination attempts, all in the shadow of an impending Olympic game. Akira inspired a cult following and had a strong influence on Western sci-fi culture that followed, including the Matrix series. Nevertheless, whereas the Japanese have suffered many data breaches prompting their government to amend the Act on the Protection of Personal Information (APPI) in 2020 [25], they are still relatively optimistic about AI. This is likely due to the familiarity of most Japanese with the long-standing use of AI and robotics in their manufacturing and health care sectors.

A good case study of this stark contrast in the vision of AI as a force for good in China and the dystopian vision in the West is the use of AI for surveillance. In China, facial recognition technology was implemented nationwide to catch criminals as well as petty misbehaviour. Most Chinese people think of such a technology, together with big data, to be beneficial in reducing crime, and in improving security and safety.

In an article titled "Facial recognition, AI and big data poised to boost Chinese public safety" in 2017 [26], Global Times reported with enthusiasm that "China now has the largest surveillance system in the world". This system is named, without irony, Skynet, after the fictional network from the Terminator films.

In another twist of the narrative of surveillance technology being a repressive tool, facial recognition, online behaviour tracking and data mining have been used successfully to discover corruption by government officials [27]. The AI system developed can detect suspicious property transfers, infrastructure construction, land acquisitions deals, and cross references between government databases to monitor and analyse the behaviour of government officials.

Meanwhile, facial recognition and surveillance technology in the West inevitably evokes public mistrust. After all, the fictional Skynet tries to prevent humans from deactivating it by launching a nuclear attack and sends robot assassins from the future. In 2001: Space Odyssey, HAL the computer also tries to kill the astronauts, after lip reading their intent to deactivate the machine. A common narrative in western science fiction is that AI will gain self-consciousness and when it does it invariably tries to overpower humans.

The reality tells another story. A study by the Carnegie Endowment for International Peace [28] has shown that while China is the largest supplier of surveillance technology to 55 countries, liberal democracies are major users of such technology. It found that "51 percent of advanced democracies deploy AI surveillance systems. In contrast, 37 percent of closed autocratic states, 41 percent of electoral autocratic/competitive autocratic states, and 41 percent of electoral democracies/illiberal democracies deploy AI surveillance technology." "Smart city platforms with a direct public security link are found in at least fifty-six of seventy-five countries with AI surveillance technology."

After centuries of natural calamities, wars, and political turmoil, the PRC entered a political reform era in 1979 which brought about 40 years of economic growth, state-sponsored entrepreneurship and private ownership, and freedom of travel. Millions of Chinese have travelled overseas as tourists, students and scholars. Since the early 1990s, there is little appetite among the Chinese population for the pondering of "existential risk". Nowadays, Chinese people flock to Hollywood blockbusters as much as any other, to consume sci-fi and fantasy entertainment starring mostly white protagonists. Only one Chinese sci-fi film, the Wandering Earth, has had the comparable budget, production quality, and the reach. As China becomes a more economically mature nation, it remains to be seen whether Chinese sci-fi will gravitate towards the cyberpunk genre, whether the Chinese public opinion will be influenced by the latter and whether they will continue to be supportive of AI in the decades to come.

4. **A Scientific Common Ground**

These gaps in both cultural representations of technology and levels of trust towards governments explains why the Chinese principles work as paternalistic guidelines where trust is not an issue, while the European principles establish the conditions for AI to be 'trustworthy' where distrust has become the norm. Despite the seemingly different, though not contradictory, approaches on AI ethics from China and the E.U., the presence of major commonalities between them points to a more promising and collaborative future in the implementation of these standards.

Much of operationalization and implementation of ethical standards in AI is in organizational governance - the *process* and *application choices* we make. When is it ethical to use facial recognition technology? How can we prevent data leakage? Is it acceptable for computer software to impersonate human agents without the knowledge of the user? How can we prevent the misuse of Deepfake AI? Should we have security auditing bodies to certify all AI products?

In addition to governance, ethical principles need to be incorporated into the *design* of AI *systems*. A wide range of AI applications are enabled by autonomous driving systems, facial recognition systems, predictive policing systems, sentiment analysis for financial market prediction, conversational AI chatbots, voice assistance systems, and other potential AI systems. A significant part of operationalizing these standards and guidelines lies in improvements and modifications to the methodology and the architecture of modern AI software *systems*.

AI systems research and development is an open and collaborative process across nations. AI algorithm designers from China, US and the EU alike are trained in a common curriculum of computer science and engineering. And STEM students around the world are educated in the shared tradition of Enlightenment and the Science Method. The Scientific Method, a paradigm of formulating hypothesis, devising empirical experiments to verify the hypothesis to arrive at a claim or thesis, has underpinned research areas from statistics, signal processing, optimization, machine learning, and pattern recognition, all forming the multidisciplinary area that is modern Artificial Intelligence today.

The Scientific Method was first adopted by China among other Enlightenment values during the May Fourth Movement in 1919. Coined the 'Chinese Enlightenment', this movement resulted in the first ever repudiation of traditional Confucian values, and it was then believed that only by adopting

Western ideas of 'Mr. Science' and 'Mr. Democracy' in place of 'Mr. Confucius' could the nation be strengthened. In the years since the third generation of Chinese leaders, the Confucian value of the 'harmonious society' is again promoted as a cultural identity of the Chinese nation. Nevertheless, 'Mr. Science' and "technological development" continue to be seen as a major engine for economic growth and livelihood improvement, hence leading to the betterment of the 'harmonious society'.

For both governance and design, two leading international standards bodies, namely the International Standards Organization (ISO) and the Institute of Electrical and Electronic Engineers (IEEE) are working on and publishing governance and best practice guidelines for the industry. Since ISO and IEEE standards lend credibility to products and services, they are widely accepted and recognized by countries all around the world. Chinese as well as EU representatives are also actively involved in these standards organizations ensuring that such standards and best practice guidelines take into account cultural norms and differences. "ISO data security standards have been widely adopted by cloud computing providers, e.g., Alibaba, Amazon, Apple, Google, Microsoft, and Tencent." [29]  The working group on IEEE Guidelines for Ethically Aligned Design [30] explored "established ethics systems, including both philosophical traditions (utilitarianism, virtue ethics, and deontological ethics) and religious and culture-based ethical systems (Buddhism, Confucianism, African Ubuntu traditions, and Japanese Shinto) and their stance on human morality in the digital age. In doing so, [... they] critique [ethical] assumptions [... and they] attempt[ed] to carry these inquiries into artificial systems' decision-making processes."

There is another reason that China acknowledges this common scientific ground in its ethical principle of "open collaboration" . Since the 1980s, hundreds of thousands of Chinese students have gone to study in the US and the EU countries, most of them in the STEM fields. American technology companies such as Microsoft, Amazon, Google, have all established research centres in the PRC where Chinese researchers are recruited to work with their counterparts in the US headquarters. Chinese graduate students in AI have one time or another worked as interns in these companies in China. A sampled study from the authors of one NeurIPS conference showed that nearly 30% of the authors received their undergraduate degrees in China, more than from any other country. Meanwhile, over 50% get their graduate degrees from the US and 16% from the EU. A significant number of Chinese AI researchers do not return to China within 5 years of completing their graduate studies in the US or EU. In recent years, top Chinese AI companies such as Tencent, Baidu, Huawei, and latecomers such as Didi and Bytedance, also have established research labs in the US and EU to attract AI talents.

Finally, the AI research community has been working increasingly on computational models of incorporating ethics into AI systems. Modern Artificial Intelligence systems, contrary to earlier generation of rule-based expert systems, are based on statistical learning from large amounts of data. This makes it nearly impossible to incorporate explicit ethical standards into these systems. Take the example of the Trolley Dilemma, where different ethical approaches would lead to different decisions of the lever puller as to whether to save the lives of multiple passengers by sacrificing one. The rationalist ethical approaches are incompatible with a statistical machine learning system. We cannot simply hardcode any rules into, say, an autonomous driving system. It is a major new research area today to design computationally feasible models of ethical principles into AI systems. Research publications on detection of model biases, toxicity, and fake news, and other areas in ethical AI systems, are subjected to the same peer-review process, and published in the same venues, as other areas of AI research. Datasets for training and metrics of measuring system performance in ethical AI are shared publicly with all researchers internationally. Since 2017,  over

50 workshops and conferences have been held in AI ethics and standards are emerging from this research. There are several annual conferences focused on AI Ethics:
- AIES: AAAI/ACM Conference on AI, Ethics, and Society (since 2018)
- ACM FAccT: Conference on Fairness, Accountability, and Transparency (since 2018)
- Annual AI for Good Global Summit (since 2017)
- The Responsible AI Forum (since 2021)
- ETHICOMP: 18th International Conference on the Ethical and Social Impacts of ICT (since 1995)

Six academic journals focused on AI ethics have been created in the past 5 years:
- AI and Ethics
- AI Ethics Journal
- The Journal of Sociotechnical Critique
- Technology and Regulation
- Artificial Intelligence – Law, Policy, & Ethics
- Ethics of AI in Context: A Multidisciplinary & Multimedia Journal

Every year, each of the top-tier machine learning and AI conferences, such as AAAI, ACL, CVPR, ICLR, ICML, IJCAI, NeurIPS, include at least one AI ethics related workshop, tutorial, theme track or topic of interest (See Appendix). First NeurIPS, then the ACL conferences, all encourage authors to include an ethical statement on the broader impact their research could have on society, including any possible negative effects.

## 5. Conclusion

In this article, we analysed and compared AI ethical guidelines from China and the E.U., from the perspective of governmental roles, of public opinion and perceptions, as well as the scientific common ground for the research and development of AI in China and the West.

The EU framework is based on the core Enlightenment values of individual freedom, equal rights and serves to protect against state abuse. The Chinese guidelines are based on the Confucian values of virtuous government, harmonious society, and targets to protect against commercial exploitation.

The EU ethical framework is also built as a dialectic system between users on one side, and AI developers and service providers on the other side. These normative rules are perceived as necessary to enable trust from users, as well as that of positive interactions between these two poles. This system is dynamic and includes effective feedback loops, allowing people to keep control and improve the system via their ability to "contest and seek effective redress against decisions made by AI systems and by the humans operating them". In other words, the transparency and explicability of AI systems are required for decisions to "be duly contested". The EU principles assumes scepticism from users and attempts to assuage such negative sentiment with protective rules.

Although the Chinese AI ethical principles seem similar to the EU in many ways (both defending fairness, robustness, transparency and interpretability, etc.), they however largely differ in the overall approach. The Chinese principles start with the assumption that Chinese citizens trust the state to guide and protect them against commercial and third-party abuses. The Chinese principles are guidelines pointing a future direction for the development of AI, rather than its limitations. The word "promote" is omnipresent to express objectives to encourage, more balanced with strict requisites than the EU's. While the EU principles require that the development of AI "must be fair", the Chinese principles state that it should "eliminate prejudices and discriminations as much as possible"; while the EU "requires processes to be transparent", China's requires to "continuously improve transparency".

Another significant difference relates to the target scope of applicability. Whereas this scope is not explicitly stated in the EU ethical framework, the Chinese principles clearly refer to the progress of the "human civilization", which necessitates the promotion of international cooperation and the creation of an international AI governance framework among stakeholders. Such a universalist ambition is both laudable for its openness and challenged as the reality of moral pluralism around the different regions of the world. Ultimately, countries could agree upon a set of common principles, it would not mean these latter would be either understood the same way by all populations (e.g. the understanding of "privacy" widely differs between France and China), nor that they would be promoted to the same extent.

Finally, the EU principles mostly refer to deontologist normative rules - mainly negative obligations, whereas the Chinese principles, stemming from Confucian values, tend to combine some strict deontologist normative rules (e.g. prohibiting evil uses and illegal activities) with softer constraints that could be satisfied on different levels (e.g. promote shared and inclusive development) and even some aspects of virtue ethics, referring to "vigilance" and "self-discipline. Finally, the Chinese principles tend to suggest directions to shape how AI should be developed and applied, whereas the EU principles aim to precisely define what it should not be allowed to do.

In conclusion, the Chinese and the EU guidelines on ethical AI are complementary and should be adopted together. They provide different levels of operational details in different dimensions and together, they reflect principles and values that are universal in their core. We propose that international standards be adopted for the global governance of AI. International standards bodies should be composed of experts in AI and ethics from different cultural backgrounds. International standards have historically been widely accepted in the industry and shaped market behaviour and social impact. Existing industry practices, commercial treaties, and international practices also encourage the adoption and enforcement of international standards by transnational actors. Organizations like the AAAI can play a vital role in advising and drafting of such standards and in their dissemination, adoption and enforcement.

## About the Authors

**Pascale Fung** is a Professor in the Department of Electronic & Computer Engineering and Department of Computer Science & Engineering at The Hong Kong University of Science & Technology (HKUST), and a visiting professor at the Central Academy of Fine Arts in Beijing. She is an elected Fellow of the Association for Computational Linguistics (ACL) for her "significant contributions towards statistical NLP, comparable corpora, and building intelligent systems that can understand and empathize with humans". She is an elected Fellow of the Institute of Electrical and Electronic Engineers (IEEE) for her "contributions to human-machine interactions", and an elected Fellow of the International Speech Communication Association for "fundamental contributions to the interdisciplinary area of spoken language human-machine interactions". She is the Director of HKUST Centre for AI Research (CAiRE), an interdisciplinary research center on top of all four schools at HKUST. She is the founding chair of the Women Faculty Association at HKUST. She is an expert on the Global Future Council, a think tank for the World Economic Forum. She represents HKUST on Partnership on AI to Benefit People and Society. She is on the Board of Governors of the IEEE Signal Processing Society. She is a member of the IEEE Working Group to develop an IEEE standard - Recommended Practice for Organizational Governance of Artificial Intelligence. Her research team has won several best and outstanding paper awards at ACL, ACL and NeurIPS workshops.

**Hubert Etienne** is a computational philosopher specializing in AI ethics and computational social science for social networks. He is a visiting scholar at the Centre for Artificial Intelligence Research (CAiRE) at the Hong Kong University of Science and Technology. He is a PhD student at the Ecole

Normale Superieure in Paris, France; a PhD resident at Facebook AI Research, where he works on the ethical implications of AI solutions for the Facebook ecosystem, and the issues they raise. He is also a research associate at the Centre for Technology & Global Affairs at Oxford University, a university lecturer on data economics at H.E.C. Paris, and a university lecturer on data ethics at Sciences Po, E.S.C.P. Europe and Institut Polytechnique.

# Appendix

# Ethics in AI

Since 2017, there have been more than 50+ events (forum, summit, conference) focusing on AI ethics. (ref: https://www.exploreaiethics.com/index/)

There are several annual conferences focused on AI Ethics: ([List](#))
- AIES: AAAI/ACM Conference on AI, Ethics, and Society (since 2018)
- ACM FAccT: [Conference on Fairness, Accountability, and Transparency](#) (since 2018)
- Annual AI for Good Global Summit (since 2017)
- The Responsible AI Forum (since 2021)
- ETHICOMP: 18th International Conference on the Ethical and Social Impacts of ICT (since 1995)

Journals focused on AI ethics in past 5 years:
- [AI and Ethics](#)
- [AI Ethics Journal](#)
- [The Journal of Sociotechnical Critique](#)
- [Technology and Regulation](#)
- [Artificial Intelligence – Law, Policy, & Ethics](#)
- [Ethics of AI in Context: A Multidisciplinary & Multimedia Journal](#)

Every year, each of top-tier ML, AI conferences include at least one AI ethics related workshop/tutorial/theme track/topic of interest.

Following is the list, but not limited to:

## 2021 (3 special themes, 6 workshops, 3 topic of interest)
- 2021 ACL: Special theme track: NLP for Social Good
  - Focused topic includes Ethics and NLP
- 2021 AAAI: Special theme track: AI for Social Impact
- 2021 IEEE ISTAS21 Technological Stewardship & Responsible Innovation

- 2021 CVPR: Workshops: 1) Ethical Considerations in Creative applications of Computer Vision, 2) Beyond Fairness: Towards a Just, Equitable, and Accountable Computer Vision
- 2021 ICLR: Workshop: ***Synthetic Data Generation: Quality, Privacy, Bias***
- 2021 IJCAI Workshop: Workshop on Deceptive AI

- 2021 NAACL: Topic of Interest: Ethics, Bias, and Fairness (first time appearing in NAACL)
- 2021 EMNLP: Topic of Interest: Ethics and NLP
- 2021 NeurIPS: Topic of Interest Social Aspects of Machine Learning (e.g., AI safety, fairness, privacy, interpretability)

## 2020 (1 special themes, 4 workshops, 1 tutorial, 2 topic of interest)
- 2020 AAAI: Special theme track: AI for Social Impact

- 2020 NeurIPS: Workshop: Fair AI in Finance
- 2020 CVPR: Workshop on Fair, Data-Efficient and Trusted Computer Vision

- 2020 ACL: <u>Tutorial</u>: Integrating Ethics into the NLP Curriculum (Introductory)

- 2020 ICML: Topic of Interest: Trustworthy Machine Learning (accountability, causality, fairness, privacy, robustness, etc.)
- 2020 NeurIPS: Topic of Interest: Social Aspects of Machine Learning (e.g., AI safety, fairness, privacy, interpretability)

- 2020 EMNLP: Ethics Panel

## 2019 (2 tutorials, 5 workshops, 1 special theme track)
- 2019 EMNLP: Tutorial: Bias and Fairness in Natural Language Processing
- 2019 NeurIPS: Tutorial: Representation Learning and Fairness
- 2019 CVPR: Workshops: 1) Fairness Accountability Transparency and Ethics in Computer Vision 2) Bias Estimation in Face Anlytics
- 2019 AAAI: Special theme track: AI for Social Impact
- 2018 CVPR: Workshop Vision with Biased or Scarce Data

Other workshops
- 2017, 2018 ACL Workshop on Ethics in Natural Language Processing (2017, 2018)
- 2019, 2020, 2021 Workshop on Gender Bias in Natural Language Processing
- 2018, 2019, 2020, 2021 Workshop on Fact Extraction and Verification (FEVER)

Major changes / highlights in ML conferences:
- 2020 Neural Information Processing Systems (NeurIPS) meeting required presenters to submit a statement on the broader impact their research could have on society, including any possible negative effects.
    - Since 2020, NeurIPS included << **Social Aspects of ML**: AI Safety; Fairness and Accountability; Privacy >> as a separate topic of interest
    - 2018 NeurIPS included fairness in one of applications
- Since 2019, AAAI has included special track "AI for social impact"
- Many NLP (NAACL, ACL, EMNLP (first in 2020) , etc) conference required to honor the ethical code set out in the ACM Code of Ethics. The consideration of the ethical impact of our research, use of data, and potential applications of our work has always been an important consideration.